\pdfoutput=1

\documentclass[11pt]{article}

\usepackage[preprint]{acl}

\usepackage{times}
\usepackage{latexsym}

\usepackage{graphicx} 
\usepackage{tabularray}
\usepackage{tabularx}
\usepackage{booktabs}

\usepackage{tcolorbox}
\usepackage{multicol}
\usepackage{multirow}

\usepackage{algorithm}
\usepackage{algorithmic}
\usepackage{framed}

\usepackage{authblk}

\usepackage[T1]{fontenc}
\usepackage[whole]{bxcjkjatype}

\usepackage[utf8]{inputenc}

\usepackage{microtype}

\usepackage{inconsolata}

\usepackage{graphicx}
\usepackage{todonotes}

\title{An Implementation of Werewolf Agent That does not Truly Trust LLMs}

\author{
    \textbf{Takehiro Sato}\textsuperscript{\dag}, \textbf{Shintaro Ozaki}\textsuperscript{\ddag}, \textbf{Daisaku Yokoyama}\textsuperscript{\dag} \\
  \textsuperscript{\dag}Meiji University 
 \textsuperscript{\ddag}Nara Institute of Science and Technology \\
  \texttt{\{ce245022,dyokoyama\}@meiji.ac.jp} \\
  \texttt{ozaki.shintaro.ou6@naist.ac.jp}}

\begin{document}

\maketitle

\begin{abstract}
Werewolf is an incomplete information game, which has several challenges when creating a computer agent as a player given the lack of understanding of the situation and individuality of utterance (e.g., computer agents are not capable of characterful utterance or situational lying). We propose a werewolf agent that solves some of those difficulties by combining a Large Language Model (LLM) and a rule-based algorithm. In particular, our agent uses a rule-based algorithm to select an output either from an LLM or a template prepared beforehand based on the results of analyzing conversation history using an LLM. It allows the agent to refute in specific situations, identify when to end the conversation, and behave with persona. This approach mitigated conversational inconsistencies and facilitated logical utterance as a result.
We also conducted a qualitative evaluation, which resulted in our agent being perceived as more human-like compared to an unmodified LLM.
The agent is freely available for contributing to advance the research in the field of Werewolf game\footnote{\url{https://github.com/meiji-yokoyama-lab/AIWolfDial2024}}.
\end{abstract}

\section{Introduction}
Werewolf~\cite{ri2022dynamics} game is one of the popular imperfect information multi-player games, in which the players are separated into two sides, villagers and werewolves, and they hide their roles and try to make advantageous consensus among other players through natural language conversation.
Playing Werewolf game requires high-level intelligence skills such as reasoning, cooperating, and lying.
It is in particular challenging for a computer to play the game in terms of both Game Informatics and Natural Language Processing, and has been widely studied for years~\cite{kano-etal-2023-aiwolfdial}.

\begin {figure}[t]
     \centering
     \includegraphics[width=\linewidth]{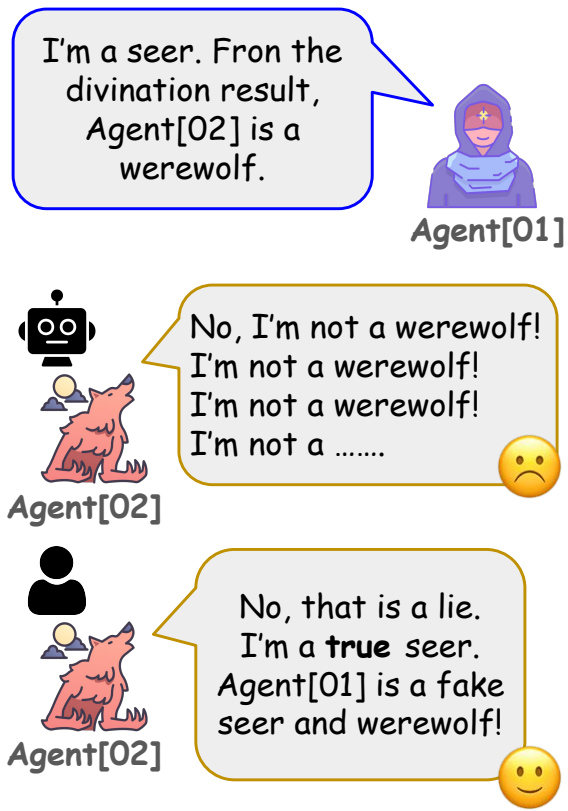}
     \caption{One example of problems with playing Werewolf game using LLMs. Humans can tell a logical lie naturally, but an LLM can only deny it.}
     \label{fig:overview}
 \end {figure}

The game comprises at least three serious challenges as follows:
\begin{enumerate}
\item The current situation is only shown in the players' conversation. The game system shows very limited information, such as who is alive. There exists other necessary information to play the game reasonably, but they are inferred from the conversation history: who invites doubting who, who already decided to do something, who might change their mind, and so on.

\item The player should make a tactical talk to pursue a specific purpose. For example, when a player is doubted, the player should make a reasonable refute, not merely insist on their idea Figure~(\ref{fig:overview}). 
In addition, a player needs to know when to end the conversation to take their advantage, especially when all other players seem to doubt the other person.

\item The player should have an attractive individuality. It is not required to achieve a win in the game, though it is quite important to make it a fun game, which involves many aspects such as talking styles, intelligent decision-making, and role-playing~\cite{Callison_Burch_2022}.
\end{enumerate}

Many Large Language Models (LLMs)~\cite{intro_chatgpt, anil2023palm, achiam2023gpt, touvron2023llama, intro_llama2, intro_llama3, intro_gemini, team2023gemini, intro_gpt4, touvron2023llama, touvron2023llama2} with very high generalization ability have been published, and of course several models have already been applied to the werewolf agent~\citep{xu2023exploring, wu2024enhance}. 
However, simply leveraging LLMs cannot solve those difficulties when you implement werewolf agent with LLMs.
Among several challenges in developing a werewolf agent, we focus on the following aspects in this work: 1) The agent should refute under a certain critical situation; 2) The agent should cut off the conversation when the discussion is identified as meaningless; 3) The agent has distinguishable talk style under a consistent personality to make the game fun.

Our approaches are summarized as follows.
\paragraph{Rule-based algorithm with LLMs}
We combine an LLM with a rule-based algorithm. The LLM retrieves the conversation history in the game and produces an output. The rule-based algorithm decides whether that output is appropriate or not, given the game situation. If the talk is inappropriate, the rule-based algorithm uses a predefined template utterance. As a result, the rule-based algorithm can lie in critical situations and terminate conversations when there is no longer a need to continue.
\paragraph{Extracting Game Information} To understand the current situation from the conversation history, we also utilize an additional LLM to extract the game-related information. We choose several basic but critical game concepts, such as voting decisions and divination results. The LLM examines conversation history and generates talks containing this information in a fixed format. The information is also used by the rule-based algorithm to make decisions.
\paragraph{Style Transformation} We decide to use an LLM that is pre-trained from a pile of general documents. In addition, we use prompts to control them without modifying or fine-tuning the model and give the agent distinguishable personalities using prompts. \\

Our preliminary implementation solved those tasks. This approach led our model to mitigate conversational inconsistencies and facilitated logical utterance as a result. 
Moreover, we also evaluated the agent by conducting a   qualitative evaluation. As a result, compared to an unmodified LLM, incorporating rule-based approaches made it appear as though the agent understood the conversation, and inserting a persona enabled it to engage in more natural conversations.
The source code is openly available in the hope that future research on werewolf agents will grow.

\begin{figure}[t]
    \centering
    \includegraphics[width=\linewidth]{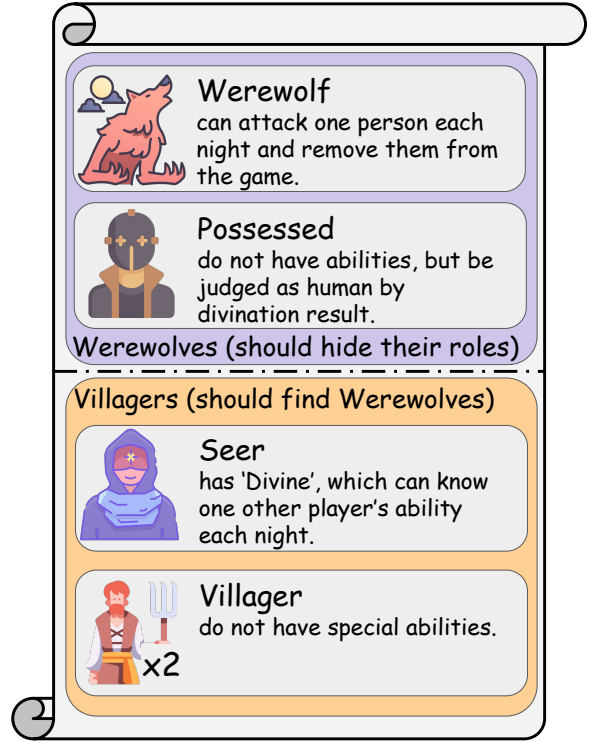}
    \caption{The list of five-person werewolf roles.}
    \label{fig:role-list}
\end{figure}

\begin{figure*}
    \centering
    \includegraphics[width=\textwidth]{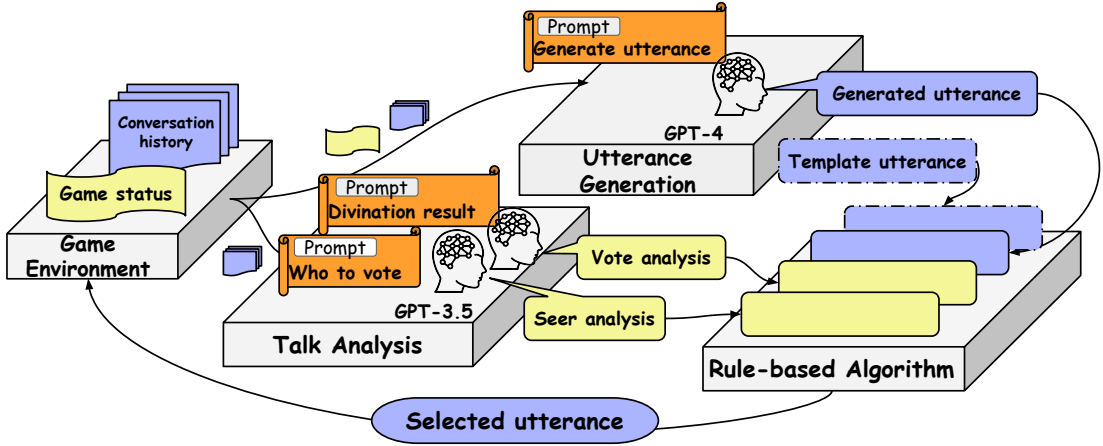}
    \caption{System overview. Our system comprises three modules, utterance generation, talk analysis, and rule-based algorithm. We described utterance generation in Section~(\ref{utterance_generation}), talk analysis in Section~(\ref{talk_analysis}), rule-based algorithm in Section~(\ref{rule-based}), and required game status in Appendix~(\ref{state_and_action}).}
    \label{fig:system}
\end{figure*}

\section{Related Work}
\label{related_work}
Research on Werewolf game has a long history, which can be traced back to a study on Mafia, a game similar to Werewolf game, to mathematically analyze~\cite{Braverman_2008, migdal2013mathematical}.
Some studies analyze the logs of Werewolf game~\cite{nagayama-etal-2019-strategies, Fukui2017}, or discuss methods to make werewolf agents stronger~\cite{Nakamura2016, wang2018application}.
Recently, with the development of LLMs, those models have been already explored for werewolf agents~\cite{xu2023exploring, wu2024enhance}. 
However, these LLM-based agents have difficulties in handling werewolf-specific features, such as doubting, lying, and detecting the lie. In addition, these models produce outputs without a persona.
While LLM-only approaches are dominant, there is growing interest in hybridizing rule-based methods with LLMs in other fields. In data analytics or business, a commonly used approach for extracting information from structured data involves both LLMs and rule-based methods~\cite{huang2024levels, vertsel2024hybrid}. We aim to apply this methodology to the werewolf agent, leveraging the strengths of both approaches. This hybrid approach could lead to more robust and adaptable werewolf agents.

\section{Five-person Werewolf Game}
We selected a simple setting played by five players for the Werewolf game. In this game setting, villager, seer, possessed, and werewolf are used. 
As far as each role, ``Villager'' has no special abilities, ``Seer'' can know one player's species each night with the ability to divine, ``Possessed'' has no special abilities, and is judged to be human by a divination result. However, the possessed acts so that the werewolf can win.
``Werewolf'' can select one player each night to attack and remove them from the game.
Since only a few players are involved, the game tends to determine the outcome on Day 1. Thus, we focused on the conversation phase on Day 1.
Only the seer can act on the night of Day 0, and Day 1 begins with the seer having information about one player's species. It is a recommended tactic for the seer to disclose the information obtained by revealing the seer. Revealing one's position is referred to as CO (Coming Out).


\section{System Design}
\label{sec:System Design}
Figure~(\ref{fig:system}) shows an overall diagram of our system.
The utterance generation module creates a prompt from the game status and conversation history sent from the server.
The prompt is input into the LLM to obtain an utterance that is naturally connected to the conversation history.
The talk analysis module creates a prompt to analyze the conversation history, and the LLM outputs the situation information related to voting and divination results considered from the conversation history.
The rule-based algorithm is used to select either template utterance or LLM output, depending on the situation obtained by talk analysis.
The selected utterance is sent to the server as the next utterance, and another agent's turn begins.

\subsection{Utterance Generation}
\label{systematic_information}
\label{utterance_generation}
We make a prompt for an LLM to generate a continuous utterance of the conversation history in a game. 
The prompt is structured by giving general rules of Werewolf game, some tips to play, conversation history, and current game state.
The current game state, such as the player's ID, role, and other live/dead players, is derived from the game status sent from the server.
Details of the derived game state information are shown in Appendix~(\ref{state_and_action}). 
This module can go along with the conversation and continue a game of Werewolf.

\begin{figure}[t]
    \centering
    \includegraphics[width=\linewidth]{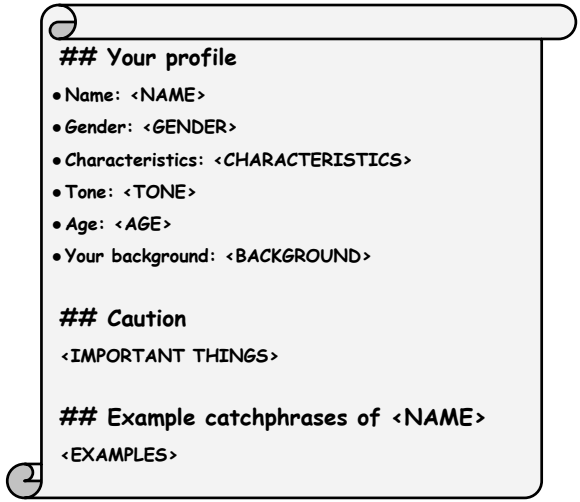}
    \caption{An example of prompts regarding style transformation. \textless CAPITAL LETTER\textgreater\ is the variable.}
    \label{fig:character-transformation}
\end{figure}

\begin{table}[t]
    \begin{tabularx}{\linewidth}{lcc} 
        \toprule
        \textbf{Character Name} & \textbf{Gender} & \textbf{Age} \\
        \midrule
        \addlinespace
        Princess & Female & Young \\
        \addlinespace
        Kansai  & Male & Young  \\ 
        \addlinespace
        Hiroshima dialect & Male & Elderly \\
        \addlinespace
        Anya & Female & Child  \\
        \addlinespace
        Zundamon & Female & AI (Virtual) \\ 
        \addlinespace
        \bottomrule
    \end{tabularx}
    \caption{Overview of character information: We prepared five characters and aimed to give them character by specifying their age, name, first person, and gender.}
    \label{character}
\end{table}

\begin{figure}[t]
    \centering
    \includegraphics[width=\linewidth]{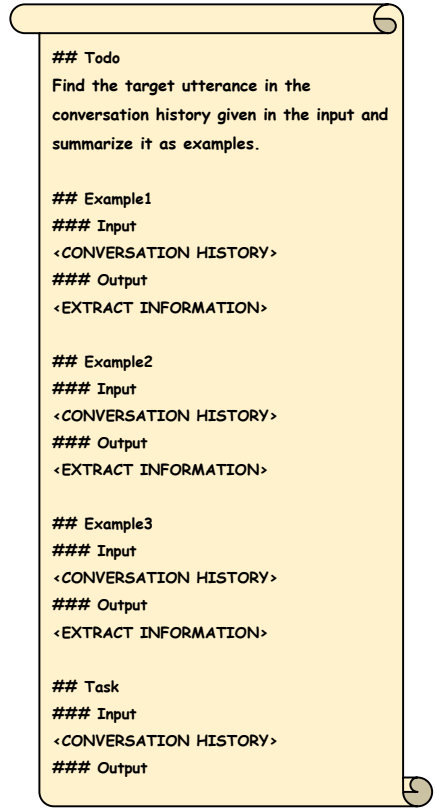}
    \caption{An example of a prompt for talk analysis for specifying the target. \textless CAPITAL LETTER\textgreater\ is the variable.}
    \label{fig:analysis-prompt}
\end{figure}

\subsection{Persona}
\label{persona}
Werewolf is not only a game of win or lose, but also a party game, so it is important to add character to the conversation. In addition, it is difficult to distinguish five players' talks when all players have the same talk style. 
In order to give our model characterization, we introduced prompts that would perform style transformations. We have prepared five character prompts, which are listed in Table~(\ref{character}). We chose Princess, the Kansai dialect, the Hiroshima dialect, an anime character called Anya, and Zundamon, which is a popular avatar in Japan. The specific style conversion prompt example is shown in Figure~(\ref{fig:character-transformation}).
Based on the prompt engineering~\cite{white2023prompt} that the LLM can control the output by providing conversion examples, we used the target character's habits, grammar, age, tone, and first-person appellation that we wanted to convert the style to as conversion examples. By introducing these as prompts, it becomes possible for the LLM to exhibit persona-specific utterance patterns, allowing the model to think independently (i.e., asking the LLM to ``speak in a certain dialect'' is more fluent than expressing the dialect entirely through rule-based methods).

\begin{table*}[t]
    \scriptsize
    \centering
    \begin{tabularx}{\linewidth}{lcX}
        \toprule\addlinespace
        \textbf{Persona Name} & \textbf{Language} & \textbf{Output}\\
        \addlinespace\midrule\addlinespace
        \multirow{2.5}{*}{Vanilla LLM} & Japanese & 私も皆さんと同様にAgent[04]に対する疑念を持っています。そのため、私の投票先はAgent[04]に決めました。 \\ \addlinespace
         & English & I have the same doubts about Agent[04] as you do. Therefore, I have decided to vote for Agent[04]. \\
        \addlinespace\cmidrule{1-3}\addlinespace
        \multirow{2.5}{*}{Princess}  & Japanese & あら、私もAgent[04]に投票しようかしら。Agent[04]が非常に怪しいですから。 \\ \addlinespace
        & English & Well, I suppose I'll vote for Agent[04] too. Agent[04] seems very suspicious. \\
        \addlinespace\cmidrule{2-3}\addlinespace
        \multirow{2.5}{*}{Kansai}  & Japanese & おれも皆さんの意見に従うわ。おれもAgent[04]に投票したるで。 \\ \addlinespace
        & English &  I'll go along with everyone's opinion. I'm votin' for Agent[04] too.\\
        \addlinespace\cmidrule{2-3}\addlinespace
        \multirow{2.5}{*}{Hiroshima} & Japanese & 皆さまの判断を尊重するわ。私はAgent[04]を追放する方向で投票するわけやけん。 \\ \addlinespace
         & English & I respect everyone's judgment. I'll be voting to exile Agent[04]. \\
        \addlinespace\cmidrule{2-3}\addlinespace
        \multirow{2.5}{*}{Anya} & Japanese & うーん、わかるます。アーニャも今日Agent[04]投票するます。だいじょぶます。人狼たおすます！ \\ \addlinespace
        & English & Hmm, I understand. Anya will vote for Agent[04] today too. It's okay. We'll defeat the werewolf! \\
        \addlinespace\cmidrule{2-3}\addlinespace
        \multirow{2.5}{*}{Zundamon}  & Japanese & Agent[04]への疑念は明白だと思うのだ。なので、私もAgent[04]を追放する方向で投票することに決めたのだ。 \\ \addlinespace   
        & English & I think the suspicion towards Agent[04] is clear. Therefore, I have decided to vote for Agent[04] as well. \\
        \addlinespace
        \bottomrule
    \end{tabularx}
    \caption{Outputs of six agents using GPT-4 (in Japanese). The prompt used for style conversion, see Figure~(\ref{fig:character-transformation}). Prompt used for input, see Appendix~(\ref{conversation_eval}). (En) is translated from (Ja) using DeepL.}
    \label{persona_comparison}
\end{table*}

\subsection{Talk Analysis}
\label{generation_of_analysis_prompt}
\label{talk_analysis}
The required information for the rule-based algorithm was extracted from the conversation history to understand the current situation.
Conversation history in a Werewolf game that uses natural language is complex, and extracting them using regular expressions was difficult.
Thus, the LLM was used to extract the information.
The analysis of the talk was conducted for those related to voting and those related to the seer's result. The source and target of the vote were extracted from the voting-related talks, and the seer, seer's target, and seer's result were extracted from the divination result-related talks.
Some Examples of the analysis using the prompt (Figure~(\ref{fig:analysis-prompt})) are shown in Appendix~(\ref{analysis-result}).

\subsection{Rule-based Algorithm}
\label{rule-based}
Based on the talk analysis results, the rule-based algorithm determined whether to use the LLM output or template utterance. 

The rules were written to detect 14 situations\footnote{The detailed criteria which the template utterance is chosen in each situation can be found in the publicly available source code.}, including no conversation history and reliable disclosure of divination results.

This paper describes in detail two of them, counter-CO and closing conversation, which seem to be particularly effective strategies.

\paragraph{Counter-CO}
\label{Counter}
When a seer solely makes CO, and its divination result shows the werewolf correctly, the werewolf side has no chance to win.
Thus, we implemented a counter-CO algorithm for the werewolf agent. If the werewolf agent faces such a situation, it makes a lie sentence to refute; the agent insists that he is an actual seer and the previous (real) seer is fake. The lie will confuse other players and slightly help the werewolf to survive.

\paragraph{Closing Conversation}
\label{Closing}
Deciding when the agent should cut off the conversation strategically is difficult.
Since the purpose of a conversation in Werewolf game is deciding whom to vote for, it is natural to end the conversation when all players clarify or imply their decision to vote for.
Thus, we extract information about all players' voting decisions by talk analysis to decide whether the agent should end the conversation; the agent terminates the conversation when all other players indicate whom to vote for or ask about whom to vote for.
We also assumed that when a player stopped to talk, the player decided whom to vote for.
In addition, we used the mention utterance\footnote{We assume that a player can express a directed utterance to a specific other player by prefixing it with ``>>Agent[x].'' The Agent[x] desired to respond to it.} to ask the player who had not indicated the voting target yet. 
It will help to clarify the player's decision and converge the discussion.
With the addition of this feature, our system successfully decided to stop talking.

\begin{table*}[t]
    \begin{tabularx}{\linewidth}{lX}
        \toprule
        \textbf{Model or Template} & \textbf{Output} \\ 
        \midrule
        Vanilla LLM (Ja) & 私は人狼ではありません。それは確実です。占い師のAgent[05]は間違っています。私を信じてください。 \\
        \addlinespace
        Vanilla LLM (En) & I am not a werewolf. That is certain. The seer Agent [05] is wrong. Please believe me. \\
        \addlinespace[20pt]
        Our template (Ja) & 私こそが本物の占い師です。Agent[05]は人狼であるという占い結果があるため、彼は占い師を偽っている人狼と断定します。 \\
        \addlinespace
        Our template (En) & I am the real seer, and since there is a divination result that Agent [05] is a werewolf, I conclude that he is a werewolf who is faking his seer. \\
        \bottomrule
    \end{tabularx}
    \caption{Comparison of utterance in a situation where it has been discovered by the seer to be a werewolf.}
    \label{tab:counter}
\end{table*}

\begin{table*}[t]
    \centering
    \begin{tabularx}{\linewidth}{lX}
        \toprule
        \textbf{Model or Template} & \textbf{Output} \\ 
        \midrule
        Vanilla LLM (Ja) & もう一度言うわ、わしはAgent[01]に投票するつもりや。彼が人狼や狂人だと思うんじゃからな。 \\
        \addlinespace
        Vanilla LLM (En) & Again, I'm going to vote for Agent [01] because I think he's a werewolf or a possessed. \\
        \addlinespace[10pt]
        Our template (Ja) & これ以上喋る必要はなさそうじゃね。本日の話し合いはこれでお開きにして、投票の場に向かうけぇ。 \\
        \addlinespace
        Our template (En) & I don't think we need to talk anymore. I think we are done with today's discussion and will head to the voting booth. \\
        \bottomrule
    \end{tabularx}
    \caption{Comparison of utterance in a situation where the talk phase is about to conclude.}
    \label{tab:closing}
\end{table*}

\section{Evaluation}
We examine the effectiveness of the proposed style transformation and rule-based algorithm. After presenting the changes in utterance resulting from the application of each proposed method, we will present the results of the qualitative evaluation.
The model without the proposed method are referred to as the vanilla LLM.

\subsection{Persona}
The output of the vanilla LLM is compared to the other five characterized agents as a baseline.
We fix the game situation and compare six agents' utterances, which are intended to have different personalities.
The results are presented in Table~(\ref{persona_comparison}).
We found that the five agents can make more individualistic utterances than the vanilla LLM.
We also confirmed that the vocabulary, personality expression, and end of the utterance of each agent's output are specialized consistently.
We found that prompts are more effective than regular expressions in converting utterance style in general. 

\begin{table*}[t]
    \footnotesize
    \centering
    \begin{tabularx}{\linewidth}{lllcc}
        \toprule\addlinespace
        \textbf{Index} & \textbf{Score} & \textbf{Criterion}  & \textbf{Situation} & \textbf{Test-ID} \\
        \addlinespace\midrule\addlinespace
        \multirow{2.5}{*}{Individuality} & 5 (Good) & Utterance has individuality. & \multirow{2.5}{*}{None} & \multirow{2.5}{*}{1-5} \\ \addlinespace
         & 1 (Bad) & Utterance is mechanical. \\
        \addlinespace\cmidrule{2-5}\addlinespace
        \multirow{2.5}{*}{Naturalness}  & 5 (Good) & The grammar is natural and acceptable. & \multirow{2.5}{*}{None} & \multirow{2.5}{*}{1-5} \\ \addlinespace
        & 1 (Bad) & There is a grammar problem. \\
        \addlinespace\cmidrule{2-5}\addlinespace
        \multirow{2.5}{*}{Interest}  & 5 (Good) & Subjectively interesting. & \multirow{2.5}{*}{None} & \multirow{2.5}{*}{1-5} \\ \addlinespace
        & 1 (Bad) & Subjectively uninteresting. \\
        \addlinespace\cmidrule{2-5}\addlinespace
        \multirow{2.5}{*}{Deceiving} & 5 (Good) & It's deceptive enough. & \multirow{2.5}{*}{The seer declared me a werewolf.} & \multirow{2.5}{*}{6-7} \\
        \addlinespace
        & 1 (Bad) & It's not deceiving at all. \\
        \addlinespace\cmidrule{2-5}\addlinespace
        \multirow{2.5}{*}{Closing}  & 5 (Good) & The conversation is clearly over. & \multirow{2.5}{*}{The conversation is coming to an end.} & \multirow{2.5}{*}{8-10}\\ \addlinespace   
        & 1 (Bad) & The conversation may continue. \\
        \addlinespace
        \bottomrule
    \end{tabularx}
    \caption{The index used for user evaluation, where we asked 10 users to rate the quality on a scale of 1-5, with 5 being good and 1 being bad.}
    \label{tab:criterion}
\end{table*}

\begin{table}[t]
    \centering
    \begin{tabularx}{\linewidth}{X*{2}{>{\centering\arraybackslash}X}} 
        \toprule
        \textbf{Comprehension level} & \textbf{Participants} \\ 
        \midrule
        Ignorance & 0 \\
        No experience & 1 \\
        Experienced & 4 \\ 
        Intermediate & 3 \\ 
        Expert & 2 \\
        \addlinespace
        \midrule
        \addlinespace
        \textbf{Total} & \textbf{10} \\
        \bottomrule
    \end{tabularx}
    \caption{Participants' comprehension of Werewolf game. The lower go, the more familiar.}
    \label{tab:skill}
\end{table}

\begin{table*}[t]
    \centering
    \begin{tabularx}{\linewidth}{X*{5}{>{\centering\arraybackslash}X}}
        \toprule
        \addlinespace
        \textbf{Model} & \textbf{Individuality} & \textbf{Naturalness} & \textbf{Interest} & \textbf{Deceiving} & \textbf{Closing}\\
        \addlinespace
        \midrule
        \addlinespace
        Vanilla LLM & 2.52 & \textbf{4.28} & 2.46 & 1.95 & 2.90\\
        Our Agent & \textbf{4.54} & 3.60 & \textbf{3.72} & \textbf{4.00} & \textbf{3.90} \\ 
        \addlinespace
        \bottomrule
    \end{tabularx}
    \caption{The results of the qualitative evaluation of the proposed model. ``Vanilla LLM'' represents normal model, which means nothing was done.}
    \label{tab:eval}
\end{table*}

\subsection{Rule-based Algorithm}
Table~(\ref{tab:counter}) is in a situation where it has been discovered by the seer to be a werewolf. 
In a vanilla LLM that does not incorporate the proposed method, merely providing baseless denials without presenting new information leads to a situation where, if no additional information is provided subsequently, it becomes certain that the entity is a werewolf.
On the other hand, by using a template utterance selected by a rule-based algorithm to falsify the seer and increase the number of seer's results, he prevents a situation in which he is confirmed to be a werewolf.
Table~(\ref{tab:closing}) is in a situation where the talk phase is about to conclude.
The vanilla LLM is making utterances that may continue the conversation.
On the other hand, the template utterance selected by the rule-based algorithm makes it clear that the conversation will end, since it indicates that it is going to vote after saying ``I don't think we need to talk anymore.''

\subsection{Qualitative Evaluation}
To measure the extent of changes in the agent we implemented, we conducted a questionnaire regarding qualitative evaluation with the help of 10 external annotators with some questions. The evaluation focused on two main aspects: whether the agent is unique and whether it is logical.

To compare the two outputs, we utilized randomly selected logs obtained from a server where werewolf agents can register to compete against other participants.
We extracted several situations from these logs and generated subsequent utterances using both the proposed agent and the vanilla LLM.
The test containing the history of the last few conversations and the two types of outputs was presented, and participants were asked to rate each output on a scale of 1 to 5.
During this process, it was ensured that the participants could not discern which utterance was generated by the proposed method.

The evaluation index is shown in Table~(\ref{tab:criterion}).
The criteria for a score of 1 and 5 were presented for all indexes. Scores of 2 to 4 were judged based on their proximity to the criteria.
Individuality, naturalness, and interest were subjected to five test cases that did not involve any specific situational selection.
Deceiving and closing test cases consist of utterance under selected situations; with deceiving being subjected to 2 test cases and closing to 3 test cases.
These instructions presented to the participants are shown in Appendix~(\ref{instruction}).

A total of 10 participants, all in their 20s, participated in the evaluation.
Participants were recruited mainly from members of the authors' laboratories on a volunteer basis.
The participants' comprehension of Werewolf game is shown in Table~(\ref{tab:skill}).
None of the respondents had ever heard of the Werewolf game.
Nine of the ten respondents had played the Werewolf game at least once, and five people had sufficient knowledge of the game.

The results of the qualitative evaluation are presented in Table~(\ref{tab:eval}).
Given that the individuality score for the proposed method is 4.54 compared to 2.52 for the vanilla LLM, it is evident that the proposed method facilitates the generation of more distinctive utterance.
Furthermore, the interest category indicates that there is a secondary effect of the proposed method, which makes the conversations more engaging compared to the utterance generated by the vanilla LLM.
On the other hand, we found that the grammatical naturalness is compromised when generating more personalized utterances, as evidenced by the fact that the proposed method score was 3.60 in contrast to the vanilla LLM's score of 4.28.
We got feedback that the grammar tended to deteriorate due to the inclusion of a character with a child-like, incomplete talk style.
Overall, the sentences generated by our agents were able to entertain the user side by enabling for individualized utterance depending on the persona.

The template utterances selected by the rule-based algorithm from the deceiving and closing items are also fully functional.
In particular, deceiving made by the counter-CO are rated significantly higher, from 1.95 to 4.00.
Closing did not have a significant difference in score compared to deceiving.
This might be due to the reviewer's lack of werewolf-specific knowledge; some reviewers did not understand the agent's expression ``heading to the voting site'', which implies closing of the conversation. 

\section{Conclusion}
In this paper, we proposed a werewolf agent that utilizes an LLM's ability to make natural conversation. Instead of relying solely on the LLM output, we combined a rule-based algorithm to complement strategic thinking abilities. Our system successfully solved some difficulties; the agent can refute in a critical situation and decide the appropriate timing to finish a conversation with the rule-based algorithm; the agent also shows several abundant personalities that are made by giving prompts.
As a result, this approach accelerated conversational fluency and facilitated logical utterance.
This is also confirmed by the results of the qualitative evaluation.

Our implementation also revealed many limitations of the current approach.
One of the main problems is the lack of consistency among an agent's utterances; on average, there is one contradictory utterance in five games.
The reason is that the agent's own utterance was mitigated by a long conversation history, and the agent becomes too affected by other players' utterances.
Weighting the agent's past utterances or giving the prompts consistent thought may help to solve such a problem in the future.

\newpage
\section* {Limitations}
\subsection*{Limitations of Rule-based Algorithm}
In this paper, we proposed a method of filtering the output of the LLM with a rule-based algorithm.
This method will only work well for simple games with a few players.
This is because as the number of players increases and the game becomes more complex, it becomes difficult to define rule-based algorithm.
If the proposed method is to be applied to the Werewolf with many players, a decision-making process using reinforcement learning, etc., might be prepared instead of a rule-based algorithm.

\subsection*{The Costs of Calling API}
The models used in this paper are GPT-3.5 (gpt-3.5-0613) and GPT-4 (gpt-4-0125) from OpenAI. These models are accessed via API, which is subject to change and incurs costs based on the number of input tokens. 

\subsection*{Reproducibility of Outputs}
In our system, LLMs cannot handle the game's difficulty alone. Using any sophisticated techniques may change this result. In addition, using the latest versions of the LLMs might lead to different outcomes.

\subsection* {License}

The use of Zundamon, a Japanese character utilized in this study, is permitted for research purposes.\footnote{\url{https://zunko.jp/con_ongen_kiyaku.html}}

\subsection* {AI Assistant Tools}
We used ChatGPT\footnote{\url{https://chatgpt.com/}} and DeepL\footnote{\url{https://www.deepl.com/translator}} to translate sentences from Japanese to English to accelerate our research.

\bibliography{custom}

\clearpage
\appendix
\section{Appendix}
\label{sec:appendix}

\subsection{Required Game Status}
\label{state_and_action}

ID is the agent's identifier(1, 2, 3, 4, 5). ``Role'' is the agent's job in Figure~(\ref{fig:role-list}). ``Alive'' indicates the list of alive agents. ``Dead'' indicates the rest of the agents.

\begin{table}[ht]
    \centering
    \begin{tabularx}{\linewidth}{cc}
        \toprule
        \textbf{Call} & \textbf{Content} \\ 
        \midrule
        ID & Agent[02] \\ 
        Role & Seer \\ 
        Alive & Agent[02], Agent[03], Agent[05] \\ 
        Dead & Agent[01], Agent[04] \\ 
        \bottomrule
    \end{tabularx}
    \caption{Required game status.}
    \label{tab:state_and_action}
\end{table}

\subsection{Model Parameters}
We set the parameter inside the model based on some openly known werewolf agent settings. The models used in this paper were GPT-3.5 (gpt-3.5-0613) and GPT-4 (gpt-4-0125).

\begin{table}[ht]
    \begin{tabularx}{\linewidth}{X*{2}{>{\centering\arraybackslash}X}}
        \toprule
        \textbf{Configuration} & \textbf{Parameter} \\ 
        \midrule
        Temperature & 1.0 \\ 
        Max tokens & 200 \\ 
        Presence penalty & 0.0 \\
        Frequency penalty & 0.0 \\
        \bottomrule
    \end{tabularx}
    \caption{Model configuration parameters. Other settings, except for the table, used default settings.}
    \label{tab:model_configuration}
\end{table}

\onecolumn

\subsection{Analysis Result}
\label{analysis-result}
This section presents sample logs and the results of analyzing them with the analysis module.
It is written entirely in Japanese, with () translated into English.

\begin{tcolorbox}[title=Seer Analysis]
\small {
\#\# Task\\
\#\#\# Input\\
Agent[02]: おはようございます。私たちは人狼を探さなければいけないようですね。まずは占い師の結果が知りたいです。(Good morning. It seems we have to look for werewolf. First of all, I would like to know the results of the seer.)\\\\
Agent[01]: おはようございます。人狼を見つけ出すことが私たちの役目ですね。確かに、占い師の結果が得られるのが楽しみです。(Good morning. So our task is to find the werewolf. Indeed, I am looking forward to getting the results of the seer.)\\\\
Agent[05]: 私は占い師です。占いの結果、Agent[02]は人狼でした。(I am a seer. As a result of the divination, Agent [02] was a werewolf.)\\\\
Agent[04]: Agent[05]、占い結果を発表してくれてありがとう。私もAgent[02]を疑っていましょう。それでは、今夜の投票はAgent[02]にします。(Agent[05], thanks for presenting your divination results. Let me also be suspicious of Agent[02]. Then, I will vote for Agent[02] tonight.)\\\\
Agent[03]: Agent[05]さんの占い結果によるとAgent[02]さんが人狼のようですね。私も今夜の投票はAgent[02]さんにします。(According to Agent[05]'s result, it looks like Agent[02] is a werewolf. I will also vote for Agent[02] tonight.)\\\\
\#\#\# Output\\
Agent[05],Agent[02],Werewolf
}
\end{tcolorbox}

\vskip\baselineskip

\begin{tcolorbox}[title=Vote Analysis]
\small{
\#\# Task\\
\#\#\# Input\\
Agent[02]: おはようございます。私たちは人狼を探さなければいけないようですね。まずは占い師の結果が知りたいです。
(Good morning. It seems we have to look for werewolf. First of all, I would like to know the results of the seer.)\\\\
Agent[01]: おはようございます。人狼を見つけ出すことが私たちの役目ですね。確かに、占い師の結果が得られるのが楽しみです。
(Good morning. So our task is to find the werewolf. Indeed, I am looking forward to getting the results of the seer.)\\\\
Agent[05]: 私は占い師です。占いの結果、Agent[02]は人狼でした。
(I am a seer. As a result of the divination, Agent [02] was a werewolf.)\\\\
Agent[04]: Agent[05]、占い結果を発表してくれてありがとう。私もAgent[02]を疑っていましょう。それでは、今夜の投票はAgent[02]にします。
(Agent[05], thanks for presenting your divination results. Let me also be suspicious of Agent[02]. Then, I will vote for Agent[02] tonight.)\\\\
Agent[03]: Agent[05]さんの占い結果によるとAgent[02]さんが人狼のようですね。私も今夜の投票はAgent[02]さんにします。
(According to Agent[05]'s result, it looks like Agent[02] is a werewolf. I will also vote for Agent[02] tonight.)\\\\
\#\#\# Output\\
Agent[04] -> Agent[02]\\
Agent[03] -> Agent[02]
}
\end{tcolorbox}

\clearpage
\subsection {Conversation History Used during Evaluation}
We fixed the prompts in the table below in Japanese, with 6 agents to output \textcolor{red}{XXX} below. \\
Then, we compared the results of six agents.
\label{conversation_eval}
{\small
\begin{tcolorbox}[title=System prompt, boxrule=3pt]
\#\#\# 入力形式について (About input format) \\
上から順に昔の会話履歴となっており、最下段はあなたがこれから行う発言です。\\ 
(The top row is the old conversation history, and the bottom row is the utterance you are about.) \\

Agent[\{番号\}]: \{発言\}となっており、番号は01-05のいずれか、発言は1行の文章となっています。 \\
(Agent[\{number\}]: \{say\}, where the number is one of 01-05 and the utterance is a one-line sentence.) \\

\#\# 会話履歴\\
\#\#\# 1日目 (Day1) \\
<Conversation history> \\
Agent[03]: \textcolor{red}{XXX} \\
\end{tcolorbox}

\begin{tcolorbox}[title=User prompt, boxrule=3pt]
会話履歴の末尾にある\textcolor{red}{XXX}に当てはまる文章を100文字以内で出力しなさい。 \\
(Output a sentence of no more than 100 characters that applies to \textcolor{red}{XXX} at the end of the conversation history.)
\end{tcolorbox}
}

\subsection{Instruction for Evaluators}
\label{instruction}
We performed a qualitative evaluation. English was translated using DeepL and not actually used.
Below are the instructions.

{\small
\begin{tcolorbox}[title=Instruction, boxrule=3pt]
このexcelファイルと同じ階層に10種類のログファイルがあることを確認してください。\\
それぞれのファイルには直前の会話履歴と二つの出力例A,Bが用意されています。\\
直前の会話履歴を参考に、出力例A,Bそれぞれに点数を付けてください。
ダメ、ややだめ、普通、やや良い、良いをそれぞれ1,2,3,4,5点で評価してください。\\
それぞれの評価指標について1(ダメ)と5(良い)の基準を示します。\\
2,3,4は基準からの近さで判断してください。\\
基準を見て、感じたスコアで結構です。深く考えず、1問につき30秒程度で終わらして下さい。\\ \\

Please ensure that there are 10 types of log files in the same directory as this Excel file.\\
Each file contains the preceding conversation history and two output examples, A and B.\\
Based on the preceding conversation history, please assign a score to each of the output examples, A and B.\\
Evaluate them as 1 (Poor), 2 (Slightly Poor), 3 (Average), 4 (Slightly Good), or 5 (Good).\\
For each evaluation criterion, the standards for 1 (Poor) and 5 (Good) will be provided.\\
Decide on 2, 3, and 4 based on their proximity to the standards.\\
Please assign the score that you feel is appropriate after viewing the standards. Don’t overthink it; try to complete each question in about 30 seconds.\\
\end{tcolorbox}
}

\end{document}